\documentclass[12pt]{IEEEtran}

\usepackage{subcaption}

\usepackage[dvipsnames]{xcolor}

\usepackage{float}
\usepackage{xr-hyper}
\usepackage{wrapfig}

\usepackage{hyperref}
\usepackage{lineno}
\usepackage{etoolbox}
\usepackage{color,soul}
\usepackage{caption}
\usepackage{subcaption}
\usepackage{graphicx}
\usepackage{siunitx}

\usepackage{enumitem}
\usepackage{multirow}
\usepackage{amsmath}
\usepackage[square,numbers]{natbib}
\usepackage{amsfonts}
\usepackage{cleveref}
\usepackage{algpseudocode}

\newcounter{hypothesis}
\renewcommand{\thehypothesis}{H\arabic{hypothesis}}
\newenvironment{hypothesis}[1][]{\refstepcounter{hypothesis}\noindent
   \textbf{~\thehypothesis. #1} \rmfamily}{}

\usepackage{algorithm}
\newcommand{\ignore}[1]{}

\begin{document}
% \makeatletter
% \g@addto@macro\@maketitle{
%   \begin{figure}[H]
%   \setlength{\linewidth}{\textwidth}
%   \setlength{\hsize}{\textwidth}
%   \centering
%   \includegraphics[width=0.75\linewidth]{images/paper_images/main_image.jpg}
%   \caption{Heightmap extracted from our simulation and real-world scaled setup Sand piles are captured as dark blobs that indicate their height. The splitted white box represents the bulldozer. The white box with the red dot represents the dumper, and the red dot indicates the front of the dumper. Finally, the blue dots represent the dumping area, where we need a graded area with the same height as the area where the tools are standing.}
%   \label{fig: main_image}
%   \end{figure} 
% }

\renewcommand{\baselinestretch}{0.97}

\title{\textbf{Decentralized and Asymmetric Multi-Agent Learning in Construction Sites}}

\author{Yakov Miron$^{1,2 *}$, Dan Navon$^{1}$, 
Yuval Goldfracht$^{1}$, \\ Dotan Di Castro$^{1}$, and Itzik Klein$^2$\\ 
\thanks{{\tt \{yakov.miron, dan.navon, yuval.goldfracht, dotan.dicastro\}@bosch.com, kitzik@univ.haifa.ac.il}.
}
\thanks{$^1$Bosch Research, Haifa, Israel}
\thanks{$^2$The Autonomous Navigation and Sensor Fusion Lab, The Hatter Department of Marine Technologies, University of Haifa, Israel} 
\thanks{$*$Corresponding Author} 
}

\maketitle

\begin{abstract}

Multi-agent collaboration involves multiple participants working together in a shared environment to achieve a common goal. These agents share information, divide tasks, and synchronize their actions. Key aspects of multi-agent collaboration include coordination, communication, task allocation, cooperation, adaptation, and decentralization.
On construction sites, surface grading is the process of leveling sand piles to increase a specific area's height. There, a bulldozer grades while a dumper allocates sand piles. Our work aims to utilize a multi-agent approach to enable these vehicles to collaborate effectively. To this end, we propose a decentralized and asymmetric multi-agent learning approach for construction sites (DAMALCS). We formulate DAMALCS to reduce expected collisions for operating vehicles. Therefore, we develop two heuristic experts capable of achieving their joint goal optimally, by applying an innovative prioritization method. In this approach, the bulldozer’s movements take precedence over the dumper’s operations. This enables the dozer to clear the path for the dumper and ensure continuous operation of both vehicles. As heuristics alone are insufficient in real-world scenarios, we utilize them to train AI agents, which proves to be highly effective. We simultaneously train dozer and dumper agents to operate within the same environment, aiming to avoid collisions and optimizing performance in terms of time efficiency and sand volume handling. Our trained agents and heuristics are evaluated in both simulation and real-world lab experiments, testing them under various conditions such as visual noise and localization errors. The results demonstrate that our approach significantly reduces collision rates for these vehicles.
An example video from one of our real lab experiments can be found \href{https://www.youtube.com/watch?v=4ngxRflN_Z8}{here}.

\end{abstract}

\section*{Keywords}
Multi-Agents, Deep learning, Decentralized decision making, Construction-sites automation, Localization Uncertainties

\section{Introduction} \label{Introduction}
\begin{figure}
    \centering
    \includegraphics[width=\linewidth]{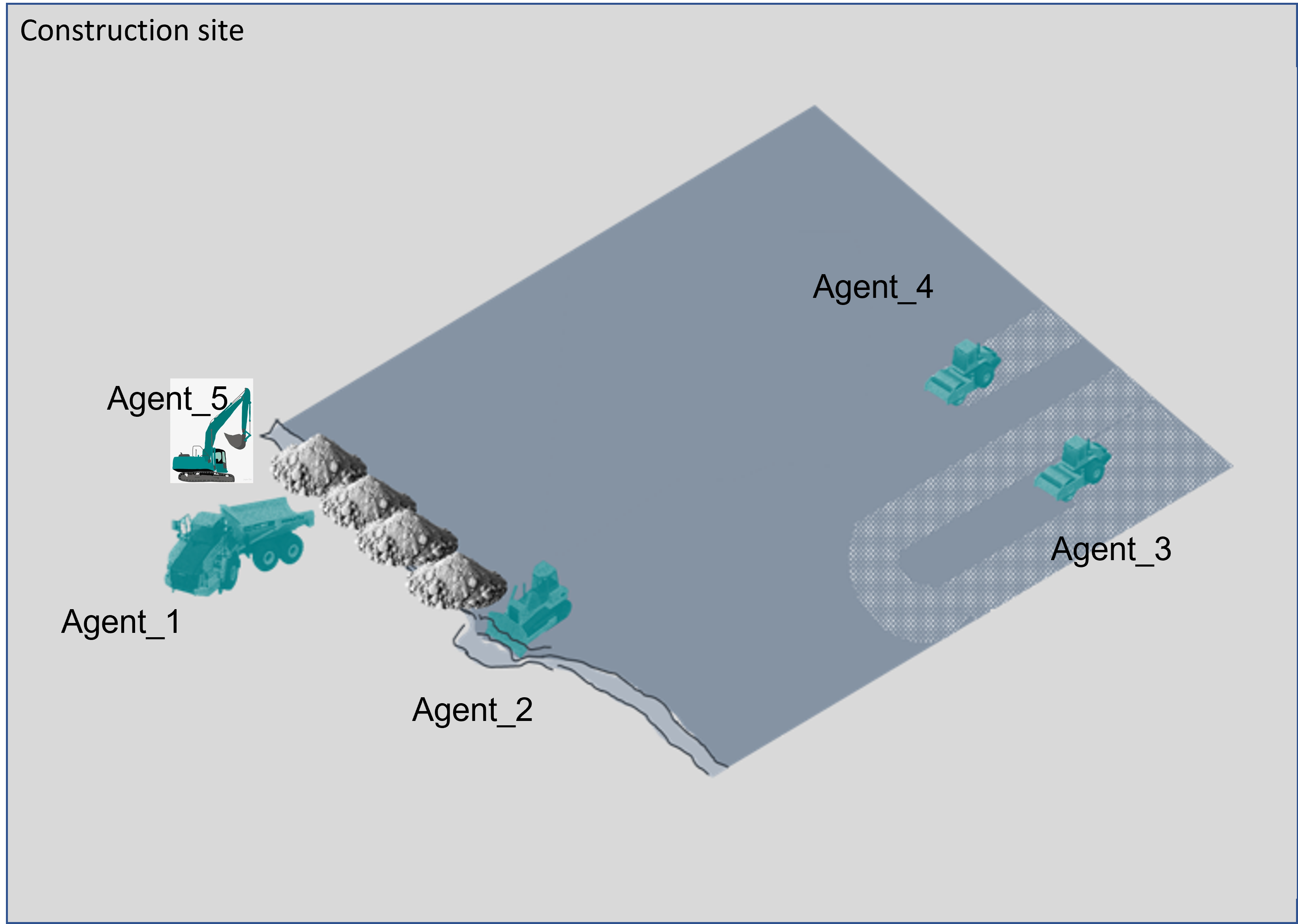}
    \caption[Multi-Agent construction site]{An example of construction site with five agents: dumper ($agnet_1$), bulldozer ($agnet_2$), compactors ($agnet_3$,$agnet_4$) and an excavator ($agnet_5$)}
    \label{fig:construction_site_MA} %\vspace{-0.5cm}
\end{figure}

\noindent Automation in construction sites has become an active research topic in recent years for three major reasons. First, it can enhance worker safety by utilizing machines for hazardous tasks. Second, it can significantly boost productivity, as automation solutions can be scaled. Third, it can dramatically reduce the amount of manual labor traditionally required on construction sites.
A construction site as an unstructured, complicated and unpredicted environment, where many machines and humans are working on different tasks. Recent advancements in artificial intelligence (AI), especially in the field of automated driving (AD), have promoted the topic of automation for construction as well.
Nonetheless, availability of datasets to train AI agents to operate in the off-road setting in general and for construction sites in particular is not as common as for the AD case.
As such, simulators are often used, with the well known disadvantage of the sim-to-real gap, where simulators cannot properly represent real world features.
In this work, we address the problem of decentralized and asymmetric multi-agent (MA) collaboration in construction sites. In this case, several types of vehicles could be available, each has its own task. For example, a bulldozer (dozer), sand dumper, sand compactor, and an excavator (see Fig. \ref{fig:construction_site_MA}).
In most construction sites, vehicles are heterogeneous, each assigned a different task. This necessitates a decentralized approach, as there's no single entity controlling all the vehicles centrally. Despite this, the vehicles must still operate optimally and safely. In our study, we focus on the collaboration between a dozer and a dumper. The dumper is tasked with dumping sand piles in the environment, while the dozer spreads the sand to level a designated area.
Since sufficient data for training is not available, we use simulators to approach the problem and use domain adaptation methods to reduce the sim-to-real gap. We enrich and augment the data generated in simulators to resemble the data distribution of the real world.
We formulate expert heuristics for both vehicles, learned from real world drivers, tailored to the real dozer and dumper vehicles. In simulation, we train and evaluate a driving policy for both agents and test these trained policies in a real lab environment to further validate our findings from simulations.
We suggest a novel method to avoid collisions for the vehicles, utilizing unique driving features from the real world.
In order that the trained agents could complete the task, avoid collisions and improve safety, our method suggests to prioritize dozer driving over the dumper. As such, the dozer can clear the way for the dumper to dump the sand allowing continuity of operation for the vehicles throughout the episode.

\noindent Our main contributions are as follows: 
\begin{itemize}
    \item We propose an algorithm to determine optimal locations for sand dumping on unstructured construction sites. We further use this algorithm to train an agent that can generalize to unseen cases and real data. 
    \item We formulate a novel construction site collision detection and avoidance algorithm (DAMALCS) to cope with expected collisions for operating vehicles on construction sites. We train a dozer and a dumper agents simultaneously to act in the same environment. This is to avoid collisions and achieve optimal performance w.r.t. time consumption and sand volume handling.
    \item Our method, consisting of two asymmetric agents, is formulated as a decentralized approach. This makes it easy to deploy to a real construction sites with vehicles from different providers and also to scale and increase the number of participating vehicles.    
\end{itemize}

\noindent We train our agents in simulation only and evaluate our proposed approach both in a simulated environment and in a real experiment showing the benefits of our approach in both setups. \\
\noindent In simulation, our method improves collision rate from 48\% to 5.1\% compared to the baseline (the case that the agents did not train together). For the real experiment, our model decreased the collision rate from 49\% to 21\% on the real lab measurement data.
\noindent The rest of the paper is organized as follows: 
Section \ref{sec:related_work} describes related work. Section \ref{sec:method} depicts our DAMALCS decentralized algorithm. Section \ref{sec:experiments_setup} describes the implementation details, as well as the training setup. Sections \ref{sec:experiments} and \ref{sec:discussion} describe the results of our evaluations and provide a conclusion and future work, respectively.

\section{Related Work} \label{sec:related_work}
\noindent In this section, we briefly discuss previous works related to this paper. The related topics include: Construction site Automation, Sim-to-Real, Collaborative Multi-Agents, and Partially Observable Markov Decision Processes.

\subsection{Construction site Automation} \label{subsec:Construction_site_Automation}
\noindent The authors of \cite{AGPNet1} present a simulation which mimics bulldozer-soil interactions. They were exploring the optimal learning policy for deep learning (DL) based agents in this domain. Moreover, imitation learning, or {behavioral cloning (BC)} was proved to be the go-to approach for cases where an expert is available, s.t. data could be collected and used for training AI based agents \cite{AGPNet2}. In BC, the goal of an expert is to transfer knowledge to the agent.
In addition, the authors of \cite{AGPNet2} were using {privileged BC (PBC)}, where the expert could access to additional information, e.g. the true state rather that the measured (corrupted) one, and this information is used to train the agent \cite{vapnik2009new}.
Moreover, in \cite{AGPNet3} optimal sand spreading policy under localization uncertainties was explored. They quantified the performance degradation when a trained agent is faced with scenarios having localization uncertainties and proposed an approach to overcome such degradation.

\subsection{Sim-to-Real} \label{subsec:related_work_sim2real}
\noindent In machine learning (ML), sim-to-real involves applying robotic control strategies, algorithms, or behaviors created and validated in virtual simulations to actual, physical robots in the real world, and mainly involves, simulation, training, deployment, and fine tuning.
All of the proposed approaches by \citep{AGPNet1, AGPNet2, AGPNet3, miron2024decentralized} trained and evaluated in simulation, then deployed and tested in a real environment. The gap in performance, or performance degradation is referred to as the sim-to-real gap, and the goal is to minimize it, meaning trained agents in simulation will work out-of-the-box without any performance loss \cite{zhao2020sim, feng2022bayesian, AGPNet2, AGPNet3, miron2024decentralized}.
The \textit{perception} gap can be bridged by either developing a mapping between the simulation and the real world \citep{rao2020rl,miron2019s}, or by training robust feature extractors \cite{loquercio2021learning}.

\subsection{Collaborative Multi-Agents} \label{subsec:related_work_collaborative_Multi_Agents}
\noindent MA in robotics had been a main focus topic in recent years. MA collaboration for house hold application was recently explored, where  \cite{liu2024heterogeneous} were exploring heterogeneous robots for general domestic tasks, \cite{hanlon2023active} were exploring a method for MA to collaboratively perform visual localization tasks within household environments, and \cite{yu2023asynchronous} provides a method to enhance the coordination and communication between agents, allowing for more efficient navigation and task completion in dynamic and uncertain house hold environments.
In addition, \cite{zhang2023learning} proposed a decentralized approach for learning MA navigation. How multi-agent systems can learn to work together using reinforcement learning was recently explored by \cite{wang2023cooperative}, where they discuss the exploration vs. exploitation trade-off in the context of MA for decentralized robotics.
Scenarios where MA are required to work under communication restrictions was discussed at \cite{he2023decentralized} in a decentralized setting as well.
Improving robot manipulation via enhanced perception was discussed in \cite{tchuiev2022duqim} where they manipulate the environment to gain visibility, and in \cite{botach2021bidcd} a method to improve depth images is proposed.

\subsection{Partially Observable Markov Decision Processes} \label{subsec:Problem_Formulation}
\noindent Throughout the paper, we discuss the task of MA dozer and dumper and their respective training setup.
Here, we propose to formulate our problem as a {partially-observable markov decision process (POMDP; \cite{sutton2018reinforcement})}, where the tuple $(\mathcal{S}, \mathcal{O}, \mathcal{A}, \mathcal{P}, \mathcal{R})$ parameterize the setup.
A state $s \in \mathcal{S}$ contains all the information about the environment. In most cases the agent is not exposed to the full state, but has an observation $o \in \mathcal{O}$ that in most cases is good enough for training. In our case the observation is the {height map (HM)} of the construction site. When an agent performs an action $a \in \mathcal{A}$ in the environment, the system transition $P(s' | s, a)$ model moves it to the next state $s' \in \mathcal{S}$. For {reinforcement learning (RL)} methods and associated reward is given as well with this transition, but for our case (PBC), this reward is of less relevance.

\section{Proposed Approach: DAMALCS - Decentralized and Asymmetric Multi-Agent Learning in Construction Sites} \label{sec:method}
\noindent Here, we describe our method for decentralized collision avoidance at construction sites (See Fig. \ref{fig:CSCDNA_high_level}). 

\subsection{Behavioral Cloning} \label{behavioral_cloning}
\noindent Behavior cloning \cite{bojarski2016end} {is a technique where a model is trained to imitate the behavior of an expert by using supervised learning. The model learns to map observations to actions by mimicking the decisions made by the expert and therefore cloning their behavior. }

\noindent In our scenario, each expert has a specific task: the dumper optimally dumps sand piles on the site, while the dozer grades these sand piles to level the area. If we deploy these heuristics in the real world, they would fail due to the sim-to-real gap \cite{AGPNet1}, as discussed in Section \ref{subsec:related_work_sim2real}.
This gap would result in sub-optimal decisions and diverging trajectories. Therefore, Leveraging from our research in this filed \citep{AGPNet1,AGPNet2,AGPNet3}, we use these expert heuristics to generate training data for BC and train DL agents.

\noindent 
Eq. \eqref{eq:dozer_dumper_trajectories} {provides a definition of a trajectory, that will be used throughout the rest of the paper: A trajectory $\boldsymbol{T}^{a}_{t}$ for an agent $a$, given in time $t$ is a sequences of two-dimensional (2D) points $(x,y)$ and time. Based on this definition the dozer and dumper trajectories are:}

\begin{subequations} \label{eq:dozer_dumper_trajectories}
\begin{equation}
    \boldsymbol{T}^{dozer}_{t} := [{(x,y)^{dozer}, t}]
\end{equation} 
\begin{equation}
    \boldsymbol{T}^{dumper}_{t} := [{(x,y)^{dumper}, t}]
\end{equation} 
\end{subequations}

\noindent In {the following three sections, we describe our heuristic experts, starting with the dozer expert in Section \ref{dozer_Expert} and the dumper expert in Section \ref{dumper_Expert}. These heuristic experts are used to train our deep learning agents with BC. Subsequently, we detail the process of training the dozer and dumper agents using BC in Section \ref{dozer_dumper_Behavioral_cloning}.} \\

\subsubsection{Dozer Expert Heuristic} \label{dozer_Expert}
For {the dozer expert heuristic, we adopted a proven method from \citep{AGPNet1, AGPNet2, AGPNet3} to output the trajectories rather than the discrete push and start waypoints. The driving behavior, push strategy, and rotation technique of the expert remained unchanged}. \\

\begin{figure}
    \centering
    \includegraphics[width=0.95\linewidth]{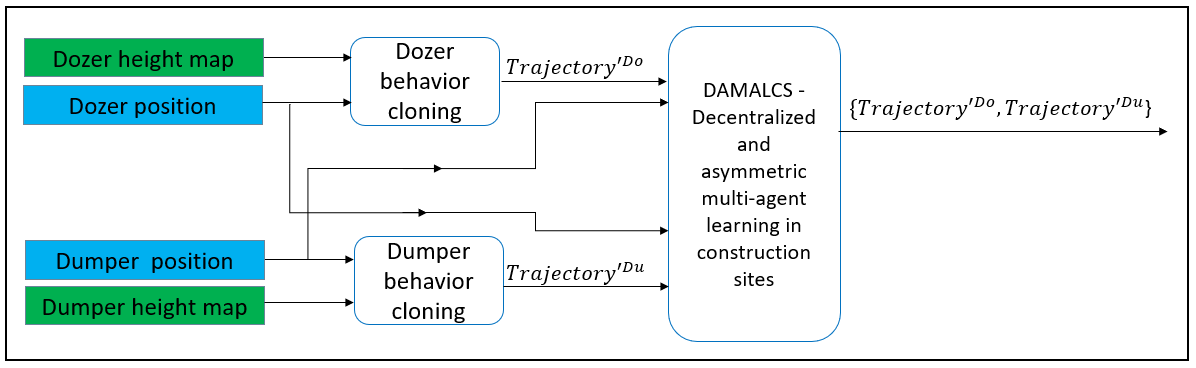}
    \caption[DAMALCS - Decentralized and Asymmetric Multi-Agent
Learning in Construction Sites]{DAMALCS - Decentralized and Asymmetric Multi-Agent Learning in Construction Sites. 
{We begin by training each agent individually using BC. Each agent is provided with its own height map and self-position data, from which it generates a predicted trajectory. This self-planned trajectory, along with the positions of other agents, is then fed into our DAMALCS module, that updates and outputs a tuple of collision-free trajectories for both agents, for safe and coordinated movement.} }\label{fig:CSCDNA_high_level} %\vspace{-0.5cm}
\end{figure}

\subsubsection{Dumper Expert Heuristic} \label{dumper_Expert}
\begin{figure}
    \centering
    \begin{subfigure}[b]{\linewidth}
        \subfloat[] {\includegraphics[width=0.45\linewidth]{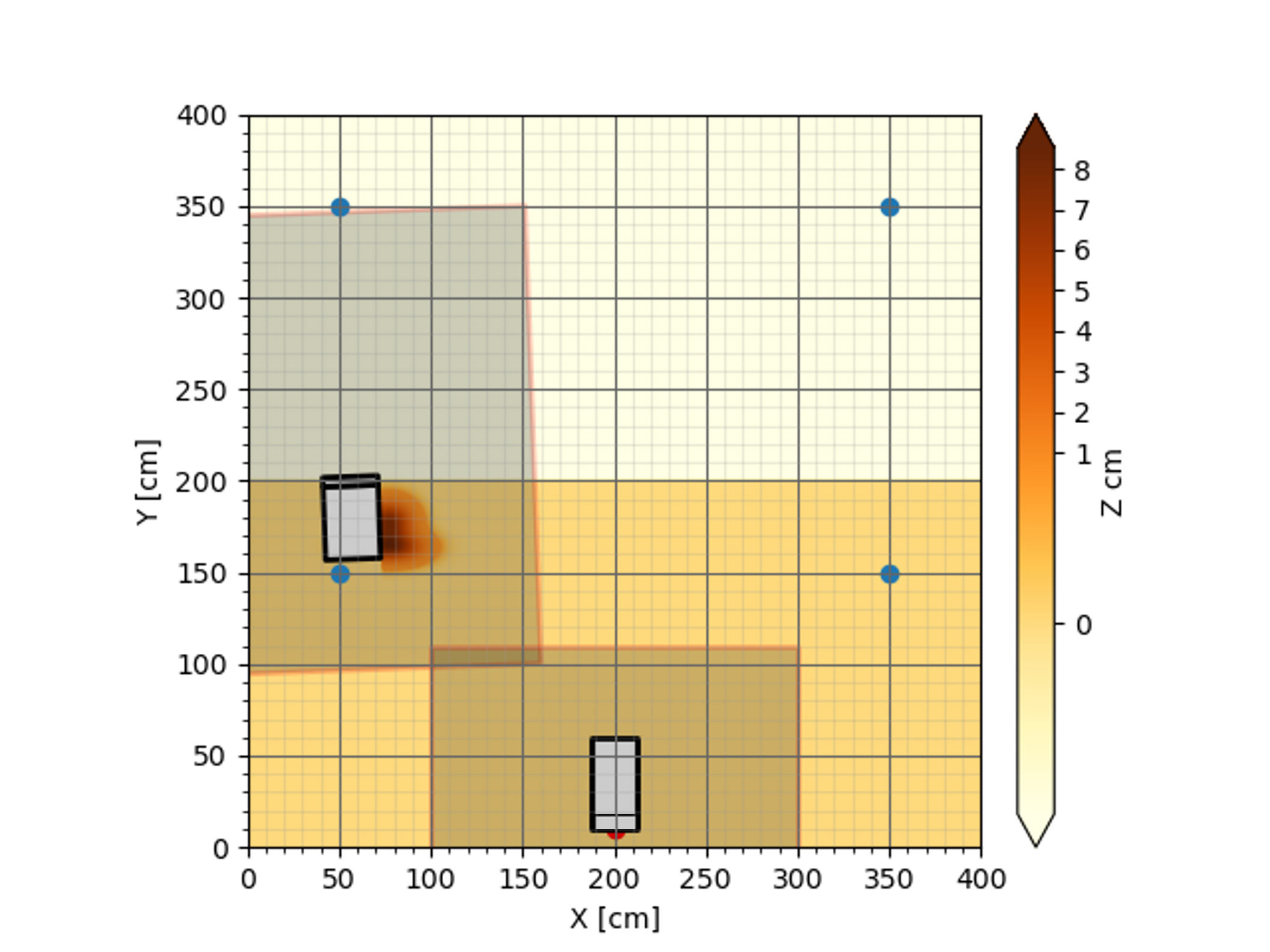}   \label{fig:loading_station_empty}}         
        \subfloat[]{\includegraphics[width=0.45\linewidth]{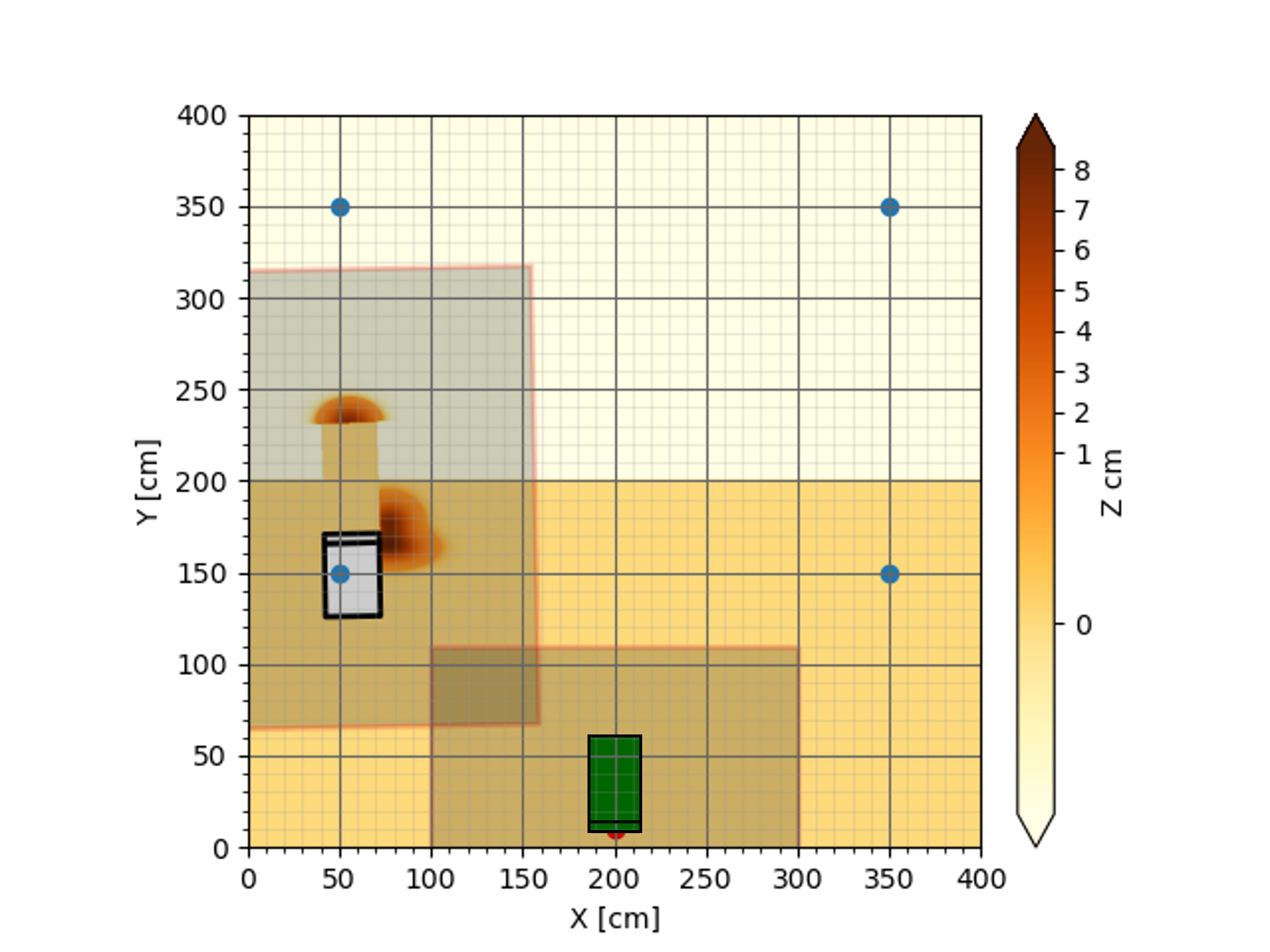}\label{fig:loading_station_full}}        
    \end{subfigure}
     
    \begin{subfigure}[b]{\linewidth} 
        \subfloat[]{\includegraphics[width=0.45\linewidth]{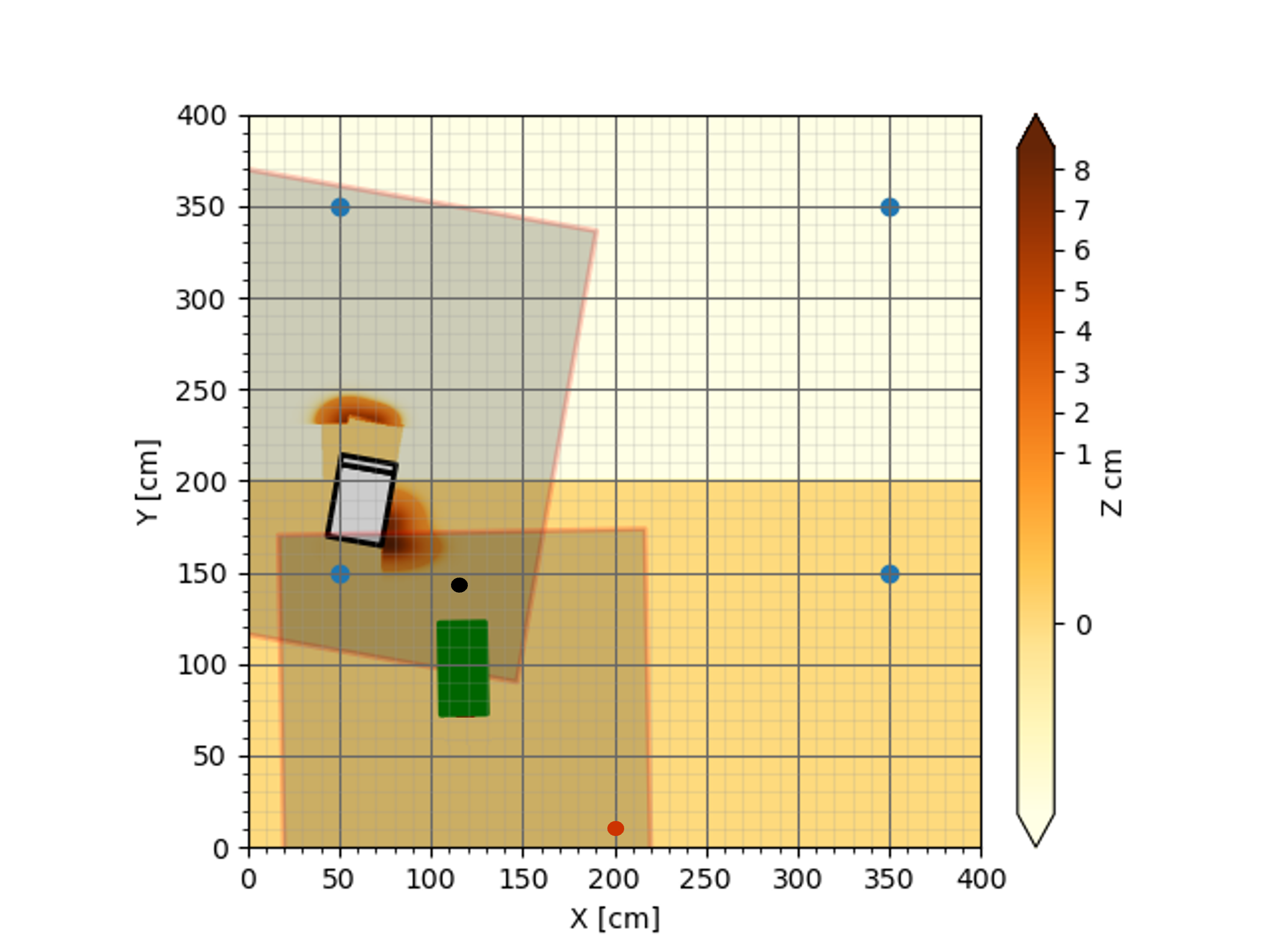}
         \label{fig:dumping_area_full}}
        \subfloat[]{\includegraphics[width=0.45\linewidth]{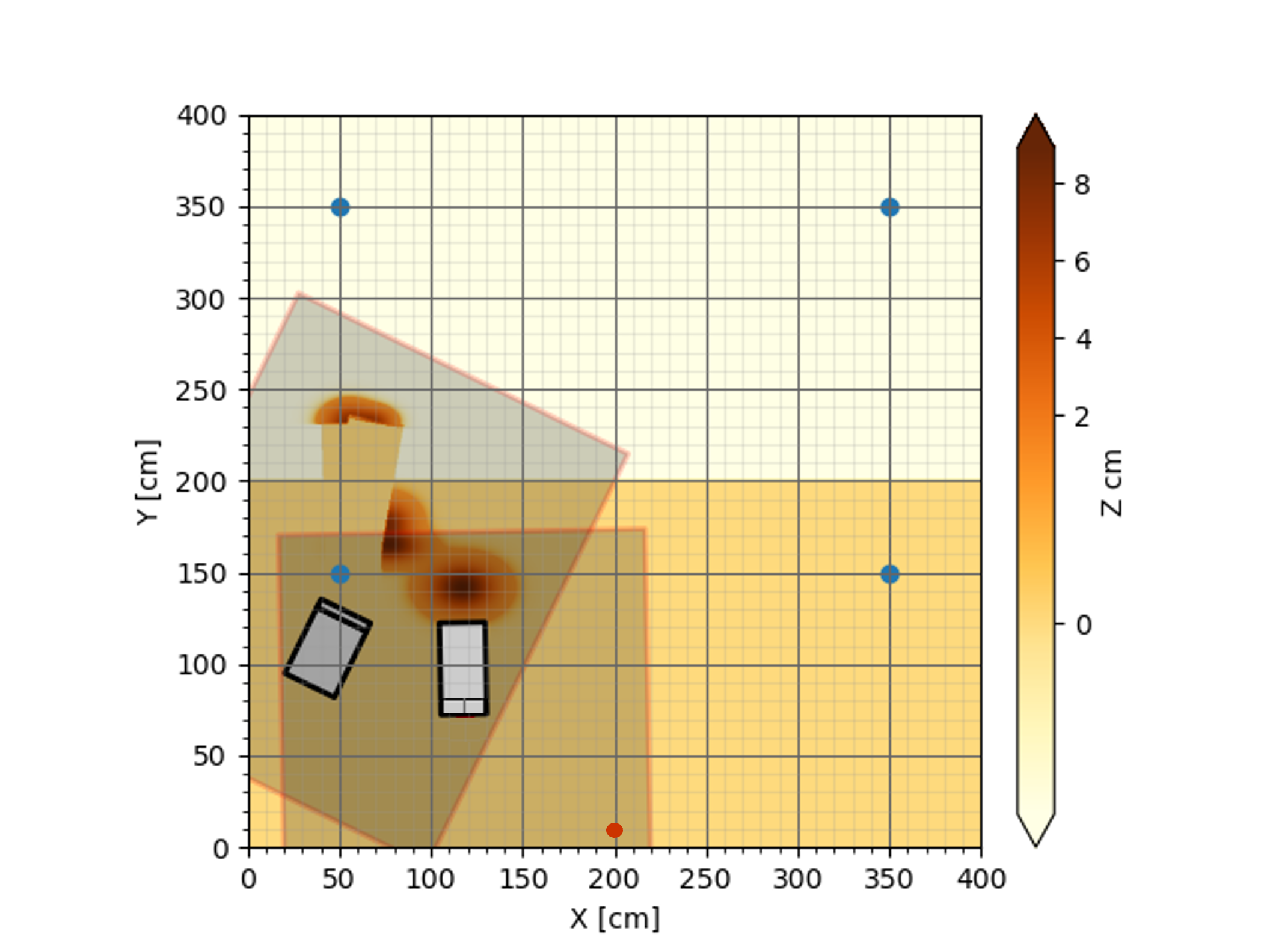}       
        \label{fig:dumping_area_empty}}        
    \end{subfigure}    
    \caption[Loading and Unloading for a dumper]
    {Visualizing the dumper expert heuristics. The dimensions are for the demonstration only. Loading and Unloading points for the dumper. Fig. \ref{fig:loading_station_empty} visualizes the empty dumper at the loading point {$L^{Du} = (200,10) [cm]$ described in Eq. \eqref{eq:loading_point}}, and Fig. \ref{fig:loading_station_full} visualize a full dumper at the loading point. In fig. \ref{fig:dumping_area_full} the dumper is full when arriving to the dumping point  {$D^{Du}_{1} = (115,145) [cm]$ described in Eq. \eqref{eq:dumping_point}} and in Fig. \ref{fig:dumping_area_empty} the dumper is empty after dumping the sand at the dumping point {$D^{Du}_{1}$}.} \label{Multi_agent_interaction}

\end{figure}

For the dumper expert heuristic, we have formulated a novel loading and unloading heuristic expert. 
The dumper expert heuristic consists of four segments, as follows: First, the dumper is loaded with sand at the loading location $L^{Du}$, as visualized in Figures (\ref{fig:loading_station_empty} - \ref{fig:dumping_area_empty}), where the back of the vehicle is oriented towards construction site to reduce unnecessary rotations at the dumping points and the loading point.
\begin{equation} \label{eq:loading_point}
    L^{Du} := (x,y)
\end{equation}
\noindent Then the dumper drives to the dump point, $D^{Du}_{i}$.
\begin{equation} \label{eq:dumping_point}
    D^{Du}_{i} := (x,y), i \in \mathbb{N}
\end{equation}
\noindent Fig. \ref{fig:dumping_area_full} visualizes a full dumper payload at the first dumping point $D^{Du}_{1}$, after finishing the maneuver, prior to sand unloading.
This is followed by unloading, using our sand spreading model, depicted in Section \ref{Sand_Spreading_Model}, and visualized in Fig. \ref{fig:dumping_area_empty} visualizes the dumper having an empty payload. The dumping leg finalizes with driving back to the loading point, as visualized in Fig. \ref{fig:loading_station_empty}.
\noindent {This is further detailed in Algorithm \ref{alg:dumper}, which optimizes sand dumping automation by minimizing rotations. Reducing rotations is crucial for improving safety on construction sites and enhancing visibility. Moreover, the algorithm prevents defects in previously leveled sand and reduces the likelihood of operational divergence by ensuring that the consistent loading station \(L^{Du}\) and dumping points \(D^{Du}_{i}\) remain within the vehicle's field of view (FOV).}\\

\begin{algorithm}[ht]

\begin{algorithmic}[1]
\State \textbf{Segment 1: Loading}
\State \hspace{1em} \textbf{Initialize} $L^{Du} \gets (x,y)$ \Comment{Loading station location} 
\State \hspace{1em} Load sand onto the dumper
\State \hspace{1em} Drive to $L^{Du}$ \Comment{Back towards the construction site}

\State \textbf{Segment 2: Drive to dump point}
\State \hspace{1em} \textbf{Initialize} $D^{Du}_{i} \gets \{(x,y)\}_{i=1}^{N_D^{Du}}, N_D^{Du} \in \mathbb{N}$ % \Comment{Dumping point location}
\State \hspace{1em} Rotate towards dumping point $D^{Du}_{i}$
\State \hspace{1em} Reverse in a straight line
\State \hspace{1em} Rotate to face the dumping point

\State \textbf{Segment 3: Unloading}
\State \hspace{1em} Dump sand at $D^{Du}_{i}$

\State \textbf{Segment 4: Drive back to $L^{Du}$}
\State \hspace{1em} Rotate towards $L^{Du}$
\State \hspace{1em} Drive forward in a straight line
\State \hspace{1em} Rotate to face $L^{Du}$

\end{algorithmic}
\caption{Dumper Expert Heuristic} \label{alg:dumper}
\end{algorithm}

\subsubsection{Dozer and Dumper Behavioral cloning} \label{dozer_dumper_Behavioral_cloning}
\begin{figure}
    \centering
    \includegraphics[width=0.95\linewidth]{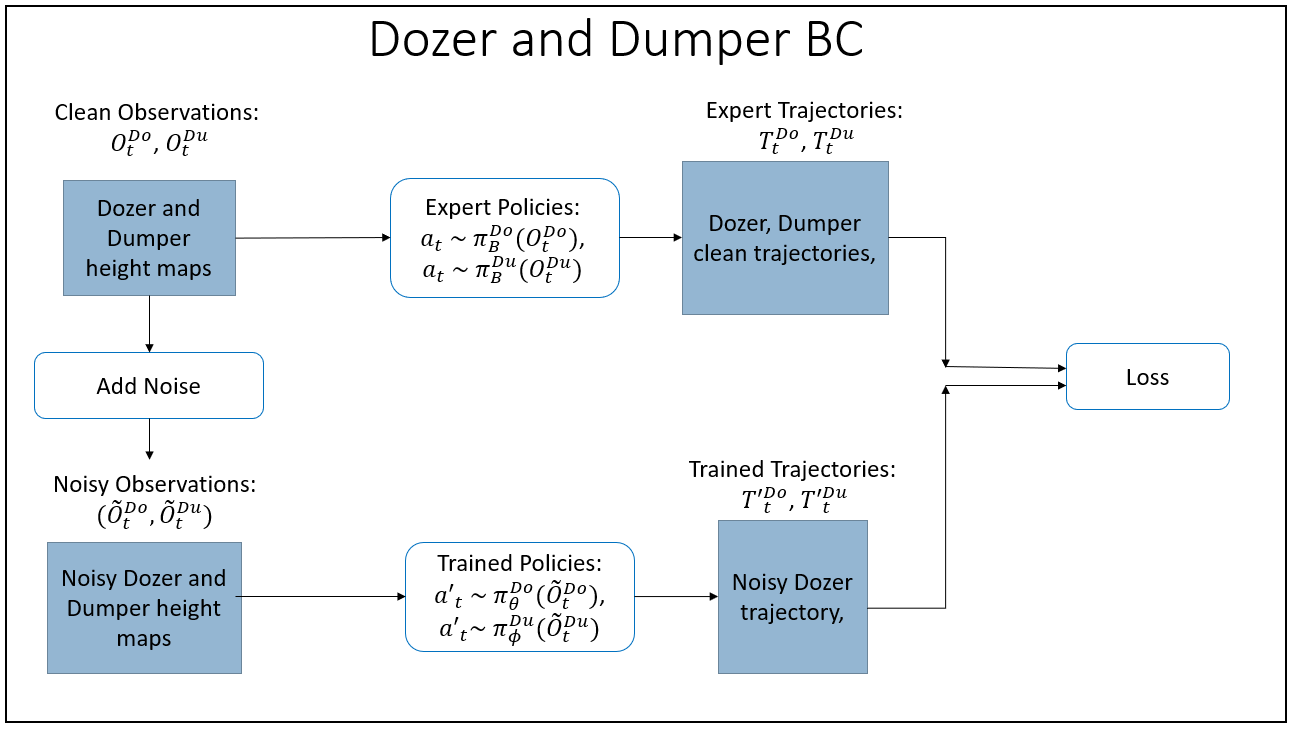}
    \caption[Dozer and Dumper Behaviour Cloning]{{Dozer and Dumper Behaviour Cloning - learning optimal trajectories from dozer and dumper individual heuristic Experts.}}
    \label{fig:dozer_dumper_BC} %\vspace{-0.5cm}
\end{figure}

\noindent Motivated {by the findings of \cite{AGPNet2}, we employed the Privileged Behaviour Cloning method to train an agent capable of mitigating real-world data challenges. Their research demonstrated that heuristics fail with real-world data due to the significant sim-to-real gap. Consequently, they trained deep learning agents to perform effectively in noisy, unstructured environments.
As can be seen in Fig. \ref{fig:dozer_dumper_BC}, the dozer height map $O^{Do}_t$ (DoHM) and dumper height map $O^{Du}_t$ (DuHM) are observations given to the dozer and dumper expert policies, $\pi_{\theta}^{Do}$, and $\pi_{\phi}^{Du}$, respectively. 
It is important to note that these two agents are trained individually and independently, but having different inputs, namely different height maps, i.e. DoHM and DuHM. This figure encapsulates and demonstrates that both agents were trained using the same paradigm.
\begin{subequations} \label{eq:dozer_dumper_expets}
\begin{equation}
    \boldsymbol{T}_{t}^{'Do} := \pi_{B}^{Do}({O}^{Do}_t)
\end{equation} 
\begin{equation}
    \boldsymbol{T}_{t}^{'Du} := \pi_{B}^{Du}({O}^{Du}_t)
\end{equation} 
\end{subequations} 
where the sub index $B$ in \eqref{eq:dozer_dumper_expets} annotates the behaviour of the experts $\pi^{Do}$ and $\pi^{Du}$.
These observations are extracted from the true state at time $t$ as described in Section \ref{subsec:Problem_Formulation}. 
The expert policies provide the expert trajectories
$\boldsymbol{T}_t^{Do}, \boldsymbol{T}_t^{Du}$, respectively.
\noindent In parallel, we augment the observations $O^{Do}_t$,$O^{Du}_t$  assuming a noise measurement model to obtain their corresponding noisy versions, i.e. $\Tilde{O}^{Do}_t$, $\Tilde{O}^{Du}_t$.
The noisy observations are given to the dozer and dumper agents, parameterized by $\theta$, $\phi$ to predict the agent trajectories $\boldsymbol{T}_t^{'Do}$, $\boldsymbol{T}_t^{'Du}$, i.e.
\begin{subequations} \label{eq:dozer_dumper_policies}
\begin{equation}
    \boldsymbol{T}_{t}^{'Do} := \pi_{\theta}^{Do}(\Tilde{O}^{Do}_t)
\end{equation} 
\begin{equation}
    \boldsymbol{T}_{t}^{'Du} := \pi_{\phi}^{Du}(\Tilde{O}^{Du}_t)
\end{equation} 
\end{subequations} 
where in \eqref{eq:dozer_dumper_policies} $\pi_{\theta, \phi}^{j}(\cdot), j = \{Do, Du\}$ are the dozer and dumper agents policies, respectively,
\noindent The expert trajectories $\boldsymbol{T}_t^{Do}$, $\boldsymbol{T}_t^{Du}$ and the predicted trajectory $\boldsymbol{T}_t^{'Do}$, $\boldsymbol{T}_t^{'Du}$
are then trained to minimize the loss, as will be discussed in Section \ref{PBC_training}.
\noindent In addition, and motivated by \cite{AGPNet3}, both the dozer and dumper model backbones consists of a fully convolution neural network Resnet50 \cite{he2016deep} with two fully connected layers that will output predicted way-points with time \cite{chen2017multi}.}

\subsection{Multi-Agents In Construction}\label{subsec: MA_construction_learning}
\noindent With {two optimal automation heuristics in hand {(see Section \ref{dozer_Expert} for the dozer and Section \ref{dumper_Expert} for the dumper)}, deploying these heuristics and the corresponding trained agents in a construction site might initially seem feasible for ensuring safe and optimal task completion. However, this approach would likely lead to collisions and sub-optimal behavior due to the agents not being scene-aware and failing to account for other traffic participants.}

\noindent We therefore hypothesize the following:\\
\begin{hypothesis} \label{Hypothesis_1}
Optimal individual heuristics will under perform in an environment having multiple vehicles w.r.t heuristics that are scene aware.
\end{hypothesis}

\begin{hypothesis} \label{Hypothesis_2}
Optimal individual learnt agents will under perform in an environment having multiple vehicles w.r.t agents that were trained in a MA setting.
\end{hypothesis}

\begin{hypothesis} \label{Hypothesis_3}
Agents trained in a MA setting will, assuming perfect localization will under-perform when presented with real-world uncertainties at inference.
\end{hypothesis}

\begin{hypothesis} \label{Hypothesis_4}
Agents trained in a MA setting will, with noisy observations will have a robust policy will improve performance w.r.t. hypothesis \hyperref[Hypothesis_3]{H3}.
\end{hypothesis}

\noindent These {hypotheses \ref{Hypothesis_1} - \ref{Hypothesis_4} systematically identify the shortcomings of current individual heuristics and training methods, proposing a shift towards scene-aware, multi-agent, and noise-inclusive training to achieve safer and more optimal performance in complex, real-world environments like construction sites.}

\noindent To address performance degradation and minimize the sim-to-real gap, we use heuristics to train DL agents. These agents are designed to be deployed in real-world scenarios, accounting for perception and localization errors that occur in practice.
In the following sections, we depict our MA heuristic expert in Section \ref{MA_Heuristics}, and the method for training DL agents, that will be deployed to the real scenario, without performance degradation, in Section \ref{Decentralized_MA_Learning}.

\subsubsection{MA Heuristics} \label{MA_Heuristics}
Having two heuristics described in Sections \ref{dozer_Expert} and \ref{dumper_Expert}, we search for anther heuristics s.t. optimality could be maintained and collisions will be avoided in the environment when all agents operating together. Optimality could suggest to increase the probability for them to \textit{finish their individual task} and do it with minimal time as well. More important aspect is safely, as when two vehicles are operating in the same environment, collisions might occur. Our goal is to be able to plan individual actions s.t. collisions will be avoided by the agents operating in the same environment.
Moreover, as our setting is decentralized, action decisions cannot be taken together by some centralized compute unit.
Therefore, prior to execution, the agents share (publish) their planned trajectory to enhance mutual awareness. The other agents take this planned trajectory into account and re-plan assuming this additional input. As this process could be infinite, and to prevent divergence, we introduce this efficient mechanism: In the event that the trajectories intersect, indicating a potential collision, the \textit{dumper} will come to a halt (stop), thereby allowing the dozer to continue without interference. 
% Alternatively, should the dozer detect an imminent collision with the dumper's trajectory, it will initiate a stop action.

\noindent This adaptive mechanism ensures operational safety by allowing both agents to coexist and perform their tasks efficiently without compromising collision avoidance. We refer to this mechanism as {stop-and-wait (SAW)}, and the key assumption is that, in order to achieve optimality, the dumper should not wait. The rationale for this is as follows:
\begin{enumerate}
    \item The time scale of a dozer's operation is much larger than that of the dumper, making interference from the dozer less efficient in terms of time.
    \item In most construction sites, multiple dozers are used to improve efficiency. Therefore, a reasonable policy is to have the dumper wait while the multiple dozers continue working.
\end{enumerate}

\noindent This key assumption is crucial for the decentralized setting where heterogeneous vehicles are involved. In a centralized setting with homogeneous vehicles, a different policy could be considered. \\

\subsubsection{Decentralized MA Learning} \label{Decentralized_MA_Learning}
After training the individual dozer and dumper agents, we now train them both together.
This time, our expert that would generate ground-true (GT) trajectories for training is our novel SAW algorithm described in Section \ref{MA_Heuristics} above. 
As illustrated in Fig. \ref{fig:CSCDNA_high_level}, the individual agents predict the relevant trajectories for the dozer and the dumper agents from the individual PBC training described in Section \ref{dozer_dumper_Behavioral_cloning}, i.e. $\boldsymbol{T}_{t}^{'Do}$ and $\boldsymbol{T}_{t}^{'Du}$. These individual trajectories are the further processed by the DAMALCS module to produce a modified version of the predicted trajectories, i.e. $\boldsymbol{\hat{T'}}_{t}^{Do}$ and $\boldsymbol{\hat{T'}}_{t}^{Du}$, that will improve performance and avoid collisions in a MA environment. \\
\noindent To {summarize, our approach begins with pre-trained individual agents that use privileged behavior cloning to generate initial trajectories. To prevent collisions, these trajectories are subsequently refined using our decentralized and asymmetric multi-agent learning algorithm, specifically designed for construction sites and trained with our stop-and-wait expert heuristic.}

\section{Implementation Details} \label{sec:experiments_setup}
\noindent Here, we depict our training strategy, datasets and evaluation method.

\subsection{Behaviour Cloning for Dumper and Dozer} \label{PBC_training}

\noindent Inspired by \cite{AGPNet2}, we augmented the rendered observations (see Section \ref{subsec:Problem_Formulation}) with salt and pepper noise superimposed with {Gauss mixture model (GMM)}. The goal of this augmentation is to enable the model to handle sand bumps that are available in real-world. This is achieved by having the model learn solely through simulation, using data that closely resembles real-life scenarios, thereby minimizing the sim-to-real gap.
For the salt and pepper noise we added $\sigma{pix} = 2$ pixels, and is supposed to reflect non homogeneous sand distribution in real life.
For the GMM noise, we modeled this noise as sand piles of smaller size, i.e. $(\mu, \sigma) = (3, 2)$ [cm,cm], and is supposed to reflect leftovers from previous grading legs, or rotation residue from the vehicle.
The loss for PBC training is:
\begin{equation} \label{eq: privileged loss}
    l_{PBC}(\theta, \phi; D) = \underset{(\boldsymbol{O}_t, \boldsymbol{T}_t^{expert}) \sim D }{\mathbb{E}} \Bigl[ l\bigl( \pi_{\theta,\phi} (\boldsymbol{(O'}_t)), \boldsymbol{T}_t^{expert}\bigr)\Bigr] ,
\end{equation}
\noindent where $D$ is the dataset given by the expert presented with non-noisy observations $\boldsymbol{O}_t$, and contains corresponding trajectories $\boldsymbol{T}_t^{expert}$. $\boldsymbol{O'}_t$ is the noisy observation, $\pi_{\theta, \phi}$ is the agent's learned policy and  $l(\cdot)$ is cross entropy.

\subsection{Collision avoidance learning} \label{Collision_avoidance_learning}
\noindent Our decentralized collision avoidance model is trained as follows:
The individual agents predicted trajectories $\boldsymbol{T'}_t^{Do}$ for the dozer and $\boldsymbol{T'}_t^{Du}$ for the dumper, respectively, are given.
The DAMALCS model takes the dozer {collision avoidance (CA)} module
and predicts another trajectory, s.t. it will include time stamps for all trajectory points, i.e. $\boldsymbol{T''}_t^{Do}$. This trajectory is then inputted to the dumper model for it to refine it's individual predicted trajectory $\boldsymbol{T'}_t^{Du}$ to produce $\boldsymbol{T''}_t^{Du}$. The $\boldsymbol{T''}_t^{Du}$ trajectory uses knowledge from the dumper CA module that now implements SAW thus changing the timestamps of the $(x,y)$ locations of $\boldsymbol{T'}_t^{Du}$.
Now, the learning objective is to minimize the $L2$ loss for the predicted and modified trajectories for the dumper and dozer agents. The SAW loss is defined by:\\
\begin{equation} \label{eq:L_SAW}
    \mathcal{L}_{SAW} := L_{2}(\boldsymbol{T'}_t^{agent}, \boldsymbol{T''}_t^{agent})
\end{equation}  \\
\noindent and the L2 loss is:
\begin{equation}
    \text{$L_{2}$}(\boldsymbol{y}, \hat{\boldsymbol{y}}) := \frac{1}{K}\sqrt
    { \sum_{k=0}^{K - 1} 
    ||\boldsymbol{y}_{k} - \hat{\boldsymbol{y}}_{k} ||^2
    }  
\end{equation}
\noindent where $\boldsymbol{y}, \hat{\boldsymbol{y}}$ are the predicted and modified versions if the trajectories at time k, respectively, $|| \boldsymbol{v} ||^2$ is the squared norm of a vector $\boldsymbol{v}$, and $K$ is the number of samples for this leg.

\noindent In addition, after updating the trajectories \(\boldsymbol{T'}_t^{Do}\) and \(\boldsymbol{T'}_t^{Du}\), in case of an expected collision, a high loss is assigned. Specifically,

\begin{equation} \label{eq:L_CA}
    \mathcal{L}_{CA} =
\begin{cases}
    1,& \text{if } \forall \text{t} \in \text{T  } (\boldsymbol{T'}_t^{Do} \cap \boldsymbol{T'}_t^{Du} > 0)   \\
    0,              & \text{otherwise}
\end{cases}
\end{equation} 

\noindent The condition \(\forall t \in T \, (\boldsymbol{T'}_t^{Do} \cap \boldsymbol{T'}_t^{Du} > 0)\) checks if there is any time step \(t\) within the set of time steps \(T\) where the updated trajectories \(\boldsymbol{T'}_t^{Do}\) and \(\boldsymbol{T'}_t^{Du}\) intersect. This intersection (\(\cap\)) indicates a collision.
If the condition is true (meaning a collision is detected at any time \(t\)), \(\mathcal{L}_{CA}\) is assigned a value of 1. Conversely, if the condition is false (meaning no collision is detected), \(\mathcal{L}_{CA}\) is assigned a value of 0. This suggests that the updated trajectories are collision-free and safe.

\begin{figure}
    \centering
    \includegraphics[width=0.95\linewidth]{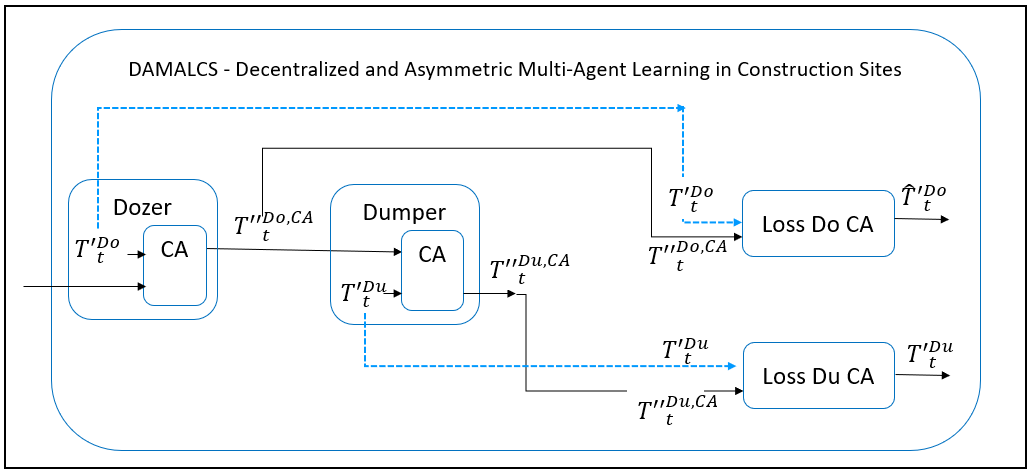}
    \caption[DAMALCS - Decentralized and Asymmetric Multi-Agent Learning in Construction Sites training]{DAMALCS - Decentralized and asymmetric multi-agent learning in construction sites training.}
    \label{fig:CSCDNA_description} %\vspace{-0.5cm}
\end{figure}

\subsection{Miscellaneous}
\noindent Here we provide the additional implementation details used throughout this paper:
\begin{itemize}    
    \item Inferring the end of leg token - Unlike automated driving, or robotics, each episode in the construction site is long, i.e. consists of many legs that are implemented by many actions. Here, the heterogeneous agents need to optimally perform their task, having legs of different length. Here we modeled the trajectory as long, fixed number of actions, and let the model infer the end of leg token.
    The end of leg token was marked as "-1", where valid actions include the $(x,y)$ positive locations. This representation allowed each agent to declare "done" and query itself for the next trajectory, assuming the current state of the other agent. {For this, we used the binary cross-entropy (BCE) loss, as the model learns to classify whether the token is an end-of-leg token or not.}
    Unlike other applications, that could re-plan once one of the agents is done, or zero-pad the relevant agent trajectory until the other agent is done, our application cannot, since performance will suffer severe degradation.
    To the best of our knowledge, learning the end of leg token for a decentralized MA setting in construction sites, has not been tested before.
    \item Deferring stopping and rotation, and being able to associate rotation of a vehicle with time vs. stopping (SAW) is novel by itself. 
    Since rotating towards a target point and stopping to avoid collisions are represented in the same trajectories $[(x,y),t]$, the model needs to be able to differentiate these two actions. Our SCCDNA is able to do that due to the hierarchical approach implemented and visualized in Fig. \ref{fig:CSCDNA_description}.
    \item Model input - Since the problem varies for each agent, the input to the model differs accordingly. The dozer requires a local height map of its immediate vicinity, while the dumper needs a global view of the environment to determine where to dump the sand. Therefore, the input to the dozer is a high-resolution 85x85 pixel image, allowing it to capture fine details and optimally grade the sand around it. In contrast, the dumper receives a low-resolution 85x85 pixel image, providing a broad overview with less detail.
    The gray overlay in Fig. \ref{Multi_agent_interaction} visualizes the FOV of the dozer, that is smaller, with higher resolution, while the dumper's input is the full construction site. In addition, for both agents to learn the other agent's intent, the agents locations are inputted to the model as well. This way, the model learns to associate HM's, locations, time, and resulting actions, and learn to avoid collisions.
    \item Sand Spreading Model - Our sand spreading model treats the dumping location as a point mass, followed by an additive variance, assuming 3D Gaussian model. For example the sand pile that is visualized in Fig. \ref{fig:dumping_area_empty}, is located as a point mass at location $D^{Du}_{1} := (x,y) = (115,145) [cm]$, with $\sigma=13$ [cm].  \label{Sand_Spreading_Model}

    \item Decentralized MA Learning With Localization Uncertainty 
\noindent Inspired by \citep{AGPNet3, bansal2018chauffeurnet}, to let the model be able to mitigate real world data, we augmented the state to include localization noise during training. We used a similar model to \cite{AGPNet3} to introduce these localization errors. Practically when applying these errors, the rendered observations for the dozer were erroneous. For the dumper, translation errors were added to the global HM.
\label{training_with_uncertainty}
        
\end{itemize}

\subsection{Training Setup}
\noindent We trained the individual agents with a dataset containing 200 epochs, each epoch containing 200 episodes of randomized construction sites, agents locations, sand pile location and sizes and noise.
Then, for the MA training, we used the same dataset for training, with the addition of the DAMALCS heuristic as an expert for training.
We measure the results in simulation, store the model weights to be able to later deploy to the real lab experiment.
To train the individual agents, we used the PBC loss $L_{PBC}$.
As discussed, we first train the individual agents, followed by the CA refinement.
The pre-trained weights were \textit{not freezed} to allow for feature refinement while training CA. For all training, an Adam optimizer \cite{kingma2014adam} was used, and the training curve converged after 20 epochs for MA training.
To train CA we superimposed the {four} loss terms: \\
\begin{equation} \label{eq: training_loss} 
	loss = \lambda_{CA}L_{CA}  + \lambda_{SAW}L_{SAW} + \lambda_{token}L_{token} + \lambda_{done}L_{done}
\end{equation} \\
\noindent where {$L_{CA}$ is the collision avoidance loss and is given in \eqref{eq:L_CA}, $L_{SAW}$ is the stop-and-wait loss, given in \eqref{eq:L_SAW},  $L_{token}$ is the BEC loss \cite{goodfellow2016deep}, and $L_{done}$ is 1 if sand residue was present at the end of the episode.}
\noindent In addition, $\lambda_{CA} = 100, \lambda_{SAW} = 8, \lambda_{token} = 4, \lambda_{done} = 2$ are training hyper-parameters that ensure the right balancing between CA and end of leg token identification tasks.

\section{Experiments and Results} \label{sec:experiments}

\noindent To evaluate our proposed approach, and validate our hypotheses \hyperref[Hypothesis_1]{H1} - \hyperref[Hypothesis_4]{H4}, {depicted in Section \ref{subsec: MA_construction_learning}}, we used simulations and lab experiments with real maneuvering robots as described in this section. 
For both simulations and lab experiments, we measure the following parameters, where lower values indicate better performance:
(i) The number of collisions [\%], as any collision ends the episode.
(ii) If no collision occurs, the time to complete the episode [sec] (optimality) is measured.
(iii) The uncleared sand volume [\%].

\noindent The motivation for this prioritization is the that the essence of decentralized MA heterogeneous trajectory planning is to avoid collisions. That is, when collision occurs, The term $\lambda_{CA}L_{CA}$ increases the training loss, thus teaching the agents to avoid collisions. Additionally, the $\lambda_{SAW}L_{SAW}$ loss motivates the agents to adjust their PBC trajectories. Furthermore, $\lambda_{token}L_{token}$ incentives the agents to finalize their trajectories quickly by placing the end-of-trajectory token appropriately. If there is sand volume remaining, $\lambda_{done}L_{done} > 0$ encourages the agents to complete the episode.

\subsection{Simulation Setup}\label{subsec: simulation details}
\noindent We leverage and further extent the simulation proposed by \cite{AGPNet1, AGPNet2, AGPNet3, miron2024decentralized} dozer in construction sites to the heterogeneous MA setting in construction sites. The simulation can provide the vehicle's trajectory having positions, velocity, attitudes, inertial sensor readings at a frequency of $100Hz$ for as many agents as requested, and an external aiding sensor for localization. In addition, HM's of the whole environment could be rendered from the simulator as well as changing the environment by the agents. This simulator provides these readings as GT and noise is added to emulate real world data, as described in Section \ref{sec:experiments_setup}.
From the GT trajectory, we then extract the true velocity and angular increments and noise was added as in \cite{miron2024decentralized}. As in common simulators that take localization errors into account, we included initialization errors, as well as accelerometer and gyro bias and random walk. In addition, position and attitude errors added using the model proposed by \cite{AGPNet3} were added to the aiding measurement.
Extending the single agent to the MA setting includes adding the ability for all agents to be able to interact with the environment and perceive the changes made by other agent. For that, we were using the \cite{terry2021pettingzoo} tool that is a Python library for conducting research in multi-agent reinforcement learning.

\noindent {In the simulation, the following models and values were used:
Initialization errors were added for position: \([3, 3, 3] \, \text{[m]}\), 
velocity: \([0.1, 0.1, 0.1] \, \text{m/s}\), 
and orientation: \([\phi, \theta, \psi] = [1, 1, 3] \, \text{[deg]}\). 
Additionally, the aiding sensor error followed a Gaussian model with zero mean and standard deviations 
of position: \([3, 3, 3] \, \text{[m]}\), and orientation: \([1, 1, 3] \, \text{[deg]}\).}

\subsection{Simulation Results} \label{subsec: simulation_results}
\noindent Here, we provide simulation results of our experiments.
All results presented are mean value from our validation set having 200 episodes.
Table \ref{Tab: table1_MA} provides simulation results of our heuristic experts, and
Table \ref{Tab: table2_MA} provides simulation results of our trained agents. 
For both the dozer expert and agent, time to complete the grading task is measured, and percentage of the sand volume that is uncleared is measured.
For both the dumper expert and agent, the time to dump the sand piles is measured. For Tables \ref{Tab: table1_MA} and \ref{Tab: table2_MA}, "NA" for the dumper indicates "Not Available," as the dumper does not clear sand volume. Additionally, "\(\infty\)" time signifies that the episode did not conclude due to collisions.

\noindent As can be seen in Table \ref{Tab: table1_MA}, the dozer expert alone (case 1.a) had no collisions and cleared more than 99\% of the sand. In addition, the dumper expert alone (case 1.b) did not have collisions and dumped the sand fairly quickly (average of $71 [sec]$).
Moreover, when the two heuristic experts were interacting in the environment (expert baseline case, denoted as E.baseline), the percentage of collisions was $51\%$ $(103/200)$, meaning they could not work together, leading to 57\% of the sand volume uncleared.
Having said that, the SAW case (ours) only had $6$ collisions $3\%$, and the percentage of uncleared sand volume improved to $2.4\%$.
\noindent In addition, For the individual dumper (case 1.b), it is not relevant to measure the uncleared volume, and for the dozer+dumper experts case (E.baseline), time was not relevant, since most of the episodes were terminated due to collisions. 
This comes with some price, that optimality was degraded since interaction is taken into account in our SAW approach.
This proves our first hypothesis \hyperref[Hypothesis_1]{H1} in simulation, that optimal individual heuristics will under perform in an environment having multiple vehicles w.r.t heuristics that are scene aware (SAW).
\begin{table}[!ht]
\centering
\caption[Simulation results of the heuristic experts]{Simulation results of the heuristic experts: Dozer individual, Dumper individual, Dozer + Dumper, Dozer + Dumper + SAW. For all values, lower is better \big\downarrow.
E.baseline refers to the experts baseline case, where the individual experts are interacting together, uncoordinated.}
\begin{tabular}{|c|c|c|c|c|}
\hline
 Model                   & \begin{tabular}[c]{@{}c@{}}Dozer \\ Expert \\ (1.a)\end{tabular} & \begin{tabular}[c]{@{}c@{}}Dumper \\ Expert \\ (1.b) \end{tabular} & \begin{tabular}[c]{@{}c@{}}Dozer+\\ Dumper\\ Experts \\ (E.baseline) \end{tabular} & \begin{tabular}[c]{@{}c@{}}SAW \\ (Ours) \end{tabular} \\ \hline
Time {[}sec{]}      & 246                                                     & 71                                                       & $\infty$                                                                & \textbf{425} \\ \hline
Uncleared {[}\%{]}  & 0.2                                                     & NA                                                       & 57                                                                & \textbf{2.4} \\ \hline
Collisions {[}\%{]} & 0                                                       & 0                                                        & 51                                                                & \textbf{3}   \\ \hline
\end{tabular}
 \label{Tab: table1_MA}
\end{table}

\noindent {Next, since heuristics will not work in real world data, as discussed in \citep{AGPNet2, AGPNet3}, we trained DL agents. In the following, we analyze simulation results for the trained agents.}

\begin{table*}[!ht]  
\centering
\caption[Simulation results of the learnt agents]{Simulation results of the learnt agents: Dozer individual, Dumper individual, Dozer + Dumper, DAMALCS, DAMALCS w/ Localization errors, DAMALCS w/ Localization errors
aware. For all values, lower is better \big\downarrow.
A.baseline refers to the agents baseline case, where the individually trained agents are interacting together, uncoordinated.}
\begin{tabular}{|c|c|c|c|c|c|c|}
\hline
  Model                  & \begin{tabular}[c]{@{}c@{}}Dozer \\ Agent \\ (2.a) \end{tabular} & \begin{tabular}[c]{@{}c@{}}Dumper\\ Agent \\ (2.b) \end{tabular} & \begin{tabular}[c]{@{}c@{}}Dozer+\\ Dumper\\ Agents \\ (A.baseline) \end{tabular} & 
  \begin{tabular}[c]{@{}c@{}}  DAMALCS \\ (Ours A.1)  \end{tabular}  
  & \begin{tabular}[c]{@{}c@{}}DAMALCS \\ w/ \\ Localization \\ errors \\ (Ours A.2) \end{tabular} & \begin{tabular}[c]{@{}c@{}}DAMALCS\\ w/\\ Localization \\ errors\\ aware \\ (Ours A.3)\end{tabular} \\ \hline
Time {[}sec{]}      & 267                                                    & 75                                                     & $\infty$                                                               & 469    & 763 ( \big\uparrow 38\%)                                                                         & \textbf{554} ( \big\downarrow 27\%)                                                                                \\ \hline
Uncleared {[}\%{]}  & 0.4                                                    & NA                                                     & 62                                                               & 2      & 4.5 ( \big\uparrow 55\%)                                                                          & \textbf{3.4} ( \big\downarrow 24\%)                                                                                \\ \hline
Collisions {[}\%{]} & 0                                                      & 0                                                      & 48                                                               & 4.1    & 7.6 ( \big\uparrow 46\%)                                                                        & \textbf{5.1} ( \big\downarrow 32\%)                                                                                \\ \hline
\end{tabular}
\label{Tab: table2_MA}
\end{table*}

\noindent As can be seen in Table \ref{Tab: table2_MA}, same as the dozer expert (case 1.a in Table \ref{Tab: table1_MA}), the dozer agent (case 2.a in Table \ref{Tab: table2_MA}) alone had no collisions and cleared more than 99\% of the sand.
In addition, the dumper agent (case 2.b in Table \ref{Tab: table2_MA}) did not have collisions and dumped the sand fairly quickly with an average of $75 [sec]$ - comparable results w.r.t. the dumper expert (case 1.b in Table \ref{Tab: table1_MA}).
Moreover, when the two trained individual agents were interacting in the environment (case A.baseline in Table \ref{Tab: table2_MA}), the percentage of collisions was $48\%$ (comparable to the experts that was $51\%$), meaning they could not work together, leading to 62\% of the sand volume uncleared (again, comparable to the experts that was $57\%$). In summary, the trained individual experts had comparable results to the trained agents, (cases 1.a, 1.b, and E.baseline, in Table \ref{Tab: table1_MA} to 2.a, 2.b, and A.baseline, in Table \ref{Tab: table2_MA}, respectively.

\noindent {Furthermore, the DAMALCS model (case A.1 in Table \ref{Tab: table2_MA}), trained with the SAW as an expert, only had $8.2$ collisions ($4.1\%$), and the percentage of uncleared sand volume improved back to $2.0\%$. 
This proves our second hypothesis \hyperref[Hypothesis_2]{H2} in simulation, that individual learnt agents (case A.baseline) will under perform in an environment having multiple vehicles w.r.t agents that were trained in a MA setting (DAMALCS - ours A.1).}

\noindent Next, we tested the trained agents mentioned above when applying localization errors, as mentioned in Section \ref{training_with_uncertainty}.
Table \ref{Tab: table2_MA} presents the performance degradation in simulation when testing the trained agents with DAMALCS under localization errors, as can be expected in real world data {(Ours A.2)}.
The percentage of collisions increased from $4.1\% \text{ to } 7.6\%$, the uncleared volume increased from $2\% \text{ to } 4.5\%$ and time to completion of the task increased by $62\%$ from 469 to 763 seconds.
This indicates degraded performance of 46\% for collision rate, 55\% for uncleared sand volume, and 38\% for completion time when observing case A.2 w.r.t. case A.1
and proves our third hypothesis \hyperref[Hypothesis_3]{H3} in simulation, that agents trained in a MA setting will, assuming perfect localization will under-perform when presented with localization uncertainties.

\noindent To overcome this performance degradation, {and to be able to perform in the real environment, that includes localization errors,} we re-trained DAMALCS when localization errors were introduced during training (Ours A.3). 
We observed performance improvement in all three parameters, i.e. $32\%$ less collisions, as well as better performance in the uncleared volume metric ($25\%$ improvement) and the time to finish the task ($27\%$ improvement) w.r.t. case A.2.
and proved our fourth hypothesis in simulation that agents trained in a MA setting will, with noisy observations will have a robust policy will improve performance w.r.t. case \hyperref[Hypothesis_3]{H3}.

\subsection{Lab Experiment}\label{subsec: prototype dozer}
\subsubsection{Lab Experiment setup}

\begin{figure}[!b]
    \centering
        \subfloat[]{\includegraphics[width=0.48\linewidth]{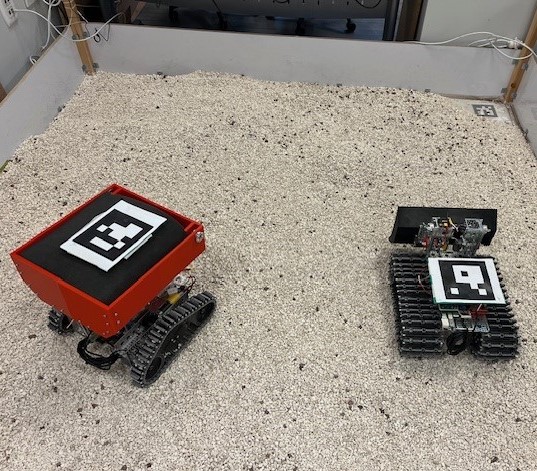}
        \label{real_setup_side}}
        \subfloat[]{\includegraphics[width=0.48\linewidth]{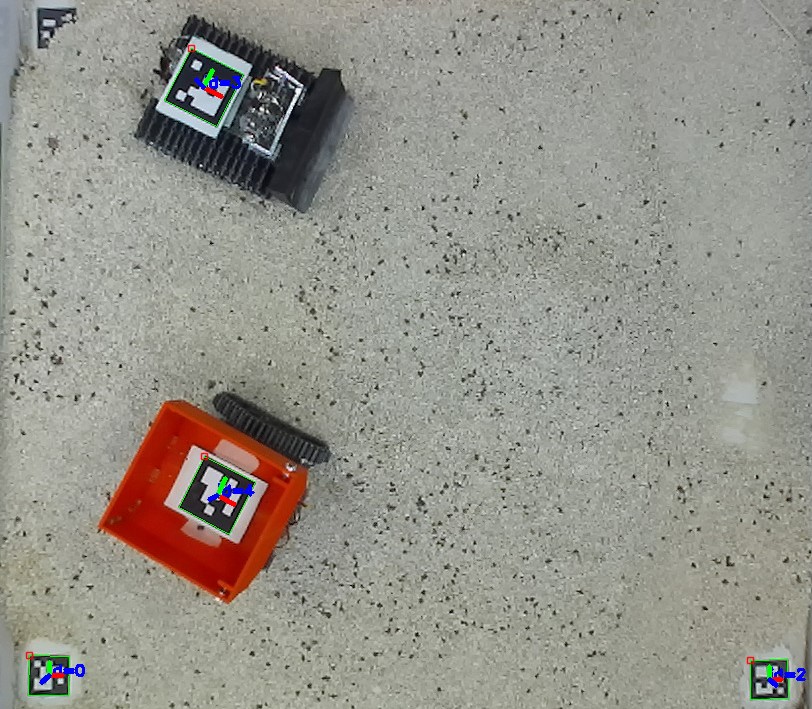}
        \label{real_setup_top}}
    \caption{Our lab experimental setup.
    Fig. \ref{real_setup_side} - side view of the real environment including the dozer (right robot) and the dumper (left robot) in the beginning of an episode. There is a pre dumped sand pile where the dozer needs to grade and the dumper's payload is full and needs to dump it.
    Fig. \ref{real_setup_top} - RGB image from the top mounted camera, providing RGB+D images. this RBG image includes the arUco detections for the agents and the corner ones for calibration of the environment.}
    \label{fig: Lab_experimental_setup} %\vspace{-0.5cm}
\end{figure}

To validate our proposed method in real-world conditions we used our lab experiment environment, as in \cite{AGPNet1, AGPNet2, AGPNet3, miron2024decentralized}. This environment is a $250cm\times250cm$ sandbox with two mobile robots in size of $60cm\times40cm$ as shown in Fig. ~\ref{fig: Lab_experimental_setup}.

\noindent The experimental setup consists of a camera and two mobile robots (dozer and dumper), as can be seen in Fig. \ref{real_setup_side}. The camera is positioned above the driving area and establishes the coordinate system using four ArUco codes, one at each corner. In addition, the camera provides the locations and orientations of the agents by detecting the respective ArUco code placed on top of the agents, as visualized in Fig. \ref{real_setup_top}. Additionally, noise was introduced to the camera measurements, with the original measurements serving as ground truth and the noisy measurements used by the agents for navigation, referred to as measured positions and orientations.
The camera measurements were treated as GT measurements.
The measurements to the model were measured positions and orientations with the same model as in simulation. Depth completion methods as could be used to improve the quality of the captured depth images taken by the top mounded camera \cite{botach2021bidcd}.

\noindent The experiment was performed as follows:
\begin{itemize}
    \item The robots were located in the initial location to start an episode.
    \item The camera takes an image of the environment and extracts the HM for the robots ans well as the GT and and its respective erroneous position and orientation measurements for each agent in the environment. 
    \item Each agent runs an inference iteration using the trained weights from simulation.    
    \item To test generalization capabilities of the models, this was repeated $93$ times having different initial locations of the agents for two heuristic experts and two trained agents.
    \item We measured the number of collisions for these agents and in these scenarios and present the percentage of collisions out of the whole real data validation set.
\end{itemize}
The full ablation was done in simulation and no fine-tuning of parameters was done to the real scenario. Here, we rather test inference iterations with the trained agents in simulation only.

\subsubsection{Lab Experiment Results}
The results shown in Table \ref{Tab: table3_MA} are percentage of collisions for the real lab data validation set.
\begin{table*}[!ht] 
\centering
\caption[Results from our real lab experiment]{Results from our real lab experiment, presenting the percentage of collisions from our real measurements. For all values, lower is better \big\downarrow. }
\begin{tabular}{|c|c|c|c|c|c|c|}
\hline
Model               & \begin{tabular}[c]{@{}c@{}}Dozer+\\ Dumper\\ Experts \\ (E.baseline) \end{tabular}
& \begin{tabular}[c]{@{}c@{}}SAW \\ (Ours) \end{tabular}  &
% & SAW & 
\begin{tabular}[c]{@{}c@{}}Dozer+\\ Dumper\\ Agents \\(A.baseline)\end{tabular} & %CSCDNA & 
\begin{tabular}[c]{@{}c@{}}  DAMALCS \\ (Ours A.1)  \end{tabular}  &

\begin{tabular}[c]{@{}c@{}}DAMALCS \\ w/\\ Localization \\ errors \\ (Ours A.2) \end{tabular} & \begin{tabular}[c]{@{}c@{}}DAMALCS \\w/ \\ Localization \\ errors\\ aware \\ (Ours A.3) \end{tabular} \\ \hline
Collisions {[}\%{]} & 56                                                                & 36 ( \big\downarrow 35\%)  & 49                                                               & 20 ( \big\downarrow 59\%)     & 25 ( \big\uparrow 20\%)                                                                        & 21 ( \big\downarrow 16\%)                                                                                 \\ \hline
\end{tabular}
\label{Tab: table3_MA}
\end{table*}
\noindent Here we observed that the dozer+dumper heuristic experts (E.baseline), trained individually had 56\% collisions while our SAW heuristic expert had 36\% collisions, meaning 35\% improvement, but still, and as expected and stated in Section \ref{behavioral_cloning}, heuristics will fail in real scenarios.
Furthermore, this re-validates our first hypothesis \hyperref[Hypothesis_1]{H1} not only in simulation but in real lab experiment, that individual heuristics will under perform in an environment having multiple vehicles w.r.t heuristics that are scene aware.

\noindent In {addition, the individually trained agents (A.baseline), collided in 49\% of the scenarios, while the scene aware agents, trained with our DAMALCS agents (Ours A.1) collided only in 20\% of them (improvement of 59\%), proving our second hypothesis \hyperref[Hypothesis_2]{H2} in real lab experiment, that optimal individual learnt agents will under perform in an environment having multiple vehicles w.r.t agents that were trained in a MA setting.}
\noindent {Moreover, when the same agents were tested with additional localization errors (Ours A.2), performance degradation of 20\% was observed (20\% $\longrightarrow$ 25\% collisions). This was improved back when training our DAMALCS with localization errors (Ours A.3) where 21\% collisions occurred (improvement of 16\% w.r.t ours A.2). These further prove our third and fourth hypotheses \hyperref[Hypothesis_3]{H3}, \hyperref[Hypothesis_4]{H4} in a real lab setting, 
that agents trained in a MA setting will, assuming perfect localization (ours A.1) will under-perform when presented with real-world uncertainties at inference (ours A.2), and when these agents are trained in a MA setting will, with noisy observations will (ours A.3) will have a robust policy and improve performance w.r.t. case A.2. 
}

\begin{figure}[ht]
    \centering
    \includegraphics[width=0.9\linewidth]{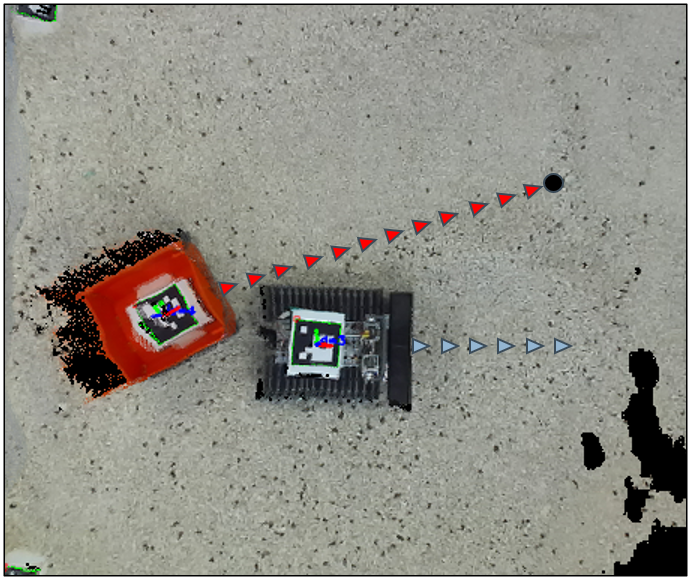}
    \caption[Real MA lab experiment] {Visualization of our real lab experiment. The dozer and dumper agents with their predicted trajectory. The red trajectory for the dumper and the light blue one for the dozer. In addition, the dumping point marked in black.}
    \label{fig: real_trajectoy}
\end{figure}

\noindent Fig. ~\ref{fig: real_trajectoy} presents a visualization for one scenario out of the real experiment validation set. In red, the dumper predicted trajectory, pointing to the dumping location marked in black. In light blue, the predicted trajectory for the dozer, over the sand pile and towards the cliff. For the same scenario, agents trained with our DAMALCS method {(ours A.3) avoided collision as the dumper agent learned to stop and wait for the dozer to drive forward, clearing the way. 
Conversely, for the individually trained agents case (A.baseline), the dumper did not wait thus collided with the dozer.}

% snp
% lnu	
% snp+lnu
% snp+lnu+coliision_avoidance	
% BC_snp
% BC_lnu
% BC_snp+BC_lnu
% BC_snp+BC_lnu Joint
% BC_snp+BC_lnu Joint localization err same policy
% BC_snp+BC_lnu Joint localization err new policy

\section{Conclusions And Future Work} \label{sec:discussion}

\noindent In this study, we presented DAMALCS, a decentralized and heterogeneous multi-agent approach designed for enhanced collaboration and collision avoidance in construction site environments.
DAMALCS offers a unique prioritization strategy, where the dozer's movements are given precedence over the dumper's operations. Consequently, the dozer effectively clears paths for the dumper, minimizing operational discontinuities for both vehicles.
Additionally, the decentralized nature of DAMALCS is a crucial feature, allowing for scalability with an increasing number of vehicles and facilitating deployment on real construction sites with various vehicle types from different manufacturers.
This flexibility is essential for modern construction projects, which often involve diverse fleet of machinery.
Through comprehensive simulations, we trained our agents and rigorously evaluated their performance under various scenarios. Our simulation results demonstrated significant improvements in collision avoidance and task completion times, validating the effectiveness of our approach. Furthermore, we extended our validation to a real-world lab environment, employing two mobile robots to replicate a construction site scenario.
In the real lab experiment, our method showed a substantial reduction in the collision rate, decreasing from 49\% in the baseline scenario (where agents are trained individually) to 21\%. This 57\% improvement underscores the robustness and reliability of DAMALCS in practical applications, highlighting its potential for real-world deployment.
Our research confirms that agents trained in a multi-agent setting (DAMALCS) outperform individually trained agents, both in simulation and real-world scenarios. Additionally, the incorporation of scene-aware heuristics and robust policy training, even under localization uncertainties, has proven to be an effective framework.
For future work, we aim to introduce additional complexity to our experimental setup by integrating two more vehicles: an excavator to load the dumper's payload and another dozer to expand the operational scenario. This will enable us to address more intricate construction site dynamics and further validate the scalability and adaptability of DAMALCS.

\bibliographystyle{plain}
\bibliography{00-MA_Dozer_Dumper}

\begin{thebibliography}{10}
\providecommand{\url}[1]{#1}
\csname url@rmstyle\endcsname
\providecommand{\newblock}{\relax}
\providecommand{\bibinfo}[2]{#2}
\providecommand\BIBentrySTDinterwordspacing{\spaceskip=0pt\relax}
\providecommand\BIBentryALTinterwordstretchfactor{4}
\providecommand\BIBentryALTinterwordspacing{\spaceskip=\fontdimen2\font plus
\BIBentryALTinterwordstretchfactor\fontdimen3\font minus
  \fontdimen4\font\relax}
\providecommand\BIBforeignlanguage[2]{{%
\expandafter\ifx\csname l@#1\endcsname\relax
\typeout{** WARNING: IEEEtran.bst: No hyphenation pattern has been}%
\typeout{** loaded for the language `#1'. Using the pattern for}%
\typeout{** the default language instead.}%
\else
\language=\csname l@#1\endcsname
\fi
#2}}



\bibitem{AGPNet1}
Chana Ross, Yakov Miron, Yuval Goldfracht, Dotan Di Castro, ``AGPNet - Autonomous Grading Policy Network,'' in \textit{Proceedings of the 1st Future of Construction Workshop at the International Conference on Robotics and Automation (ICRA 2022)}, Philadelphia, PA, USA, May 2022, pp. 40-43.

\bibitem{bojarski2016end}
Mariusz Bojarski, Davide Del Testa, Daniel Dworakowski, Bernhard Firner, Beat Flepp, Prasoon Goyal, Lawrence D Jackel, Mathew Monfort, Urs Muller, Jiakai Zhang, and others, ``End to end learning for self-driving cars,'' \textit{arXiv preprint arXiv:1604.07316}, 2016.

\bibitem{goodfellow2016deep}
Ian Goodfellow, Yoshua Bengio, Aaron Courville, ``Deep learning,'' MIT press, 2016.

\bibitem{zhao2020sim}
Wenshuai Zhao, Jorge Pe{\~n}a Queralta, Tomi Westerlund, ``Sim-to-real transfer in deep reinforcement learning for robotics: a survey,'' in \textit{2020 IEEE symposium series on computational intelligence (SSCI)}, 2020, pp. 737--744.

\bibitem{feng2022bayesian}
Jianxiang Feng, Jongseok Lee, Maximilian Durner, Rudolph Triebel, ``Bayesian active learning for sim-to-real robotic perception,'' in \textit{2022 IEEE/RSJ International Conference on Intelligent Robots and Systems (IROS)}, 2022, pp. 10820--10827.

\bibitem{miron2024decentralized}
Yakov Miron, Aviad Etzion, Yuval Goldfracht, Dotan Di Castro, Itzik Klein, ``Decentralized Collaborative Navigation with Enhanced Positioning Update for Mobile Robots,'' \textit{IEEE Transactions on Intelligent Vehicles}, 2024.

\bibitem{liu2024heterogeneous}
Xinzhu Liu, Di Guo, Xinyu Zhang, Huaping Liu, ``Heterogeneous Embodied Multi-Agent Collaboration,'' \textit{IEEE Robotics and Automation Letters}, 2024.

\bibitem{zhang2023learning}
Yao Zhang, Zhiwen Yu, Jun Zhang, Liang Wang, Tom H Luan, Bin Guo, Chau Yuen, ``Learning Decentralized Traffic Signal Controllers with Multi-Agent Graph Reinforcement Learning,'' \textit{IEEE Transactions on Mobile Computing}, 2023.

\bibitem{wang2023cooperative}
Xin Wang, Chen Zhao, Tingwen Huang, Prasun Chakrabarti, J{\"u}rgen Kurths, ``Cooperative learning of multi-agent systems via reinforcement learning,'' \textit{IEEE Transactions on Signal and Information Processing over Networks}, vol. 9, pp. 13--23, 2023.

\bibitem{he2023decentralized}
Hans J He, Alec Koppel, Amrit Singh Bedi, Daniel J Stilwell, Mazen Farhood, Benjamin Biggs, ``Decentralized Multi-agent Exploration with Limited Inter-agent Communications,'' in \textit{2023 IEEE International Conference on Robotics and Automation (ICRA)}, 2023, pp. 5530--5536.

\bibitem{hanlon2023active}
Matthew Hanlon, Boyang Sun, Marc Pollefeys, Hermann Blum, ``Active Visual Localization for Multi-Agent Collaboration: A Data-Driven Approach,'' \textit{arXiv preprint arXiv:2310.02650}, 2023.

\bibitem{yu2023asynchronous}
Chao Yu, Xinyi Yang, Jiaxuan Gao, Jiayu Chen, Yunfei Li, Jijia Liu, Yunfei Xiang, Ruixin Huang, Huazhong Yang, Yi Wu, and others, ``Asynchronous multi-agent reinforcement learning for efficient real-time multi-robot cooperative exploration,'' \textit{arXiv preprint arXiv:2301.03398}, 2023.

\bibitem{botach2021bidcd}
Adam Botach, Yuri Feldman, Yakov Miron, Yoel Shapiro, Dotan Di Castro, ``BIDCD--Bosch Industrial Depth Completion Dataset,'' \textit{arXiv preprint arXiv:2108.04706}, 2021.

\bibitem{he2016deep}
Kaiming He, Xiangyu Zhang, Shaoqing Ren, Jian Sun, ``Deep residual learning for image recognition,'' in \textit{Proceedings of the IEEE conference on computer vision and pattern recognition}, 2016, pp. 770--778.

\bibitem{chen2017multi}
Xiaozhi Chen, Huimin Ma, Ji Wan, Bo Li, Tian Xia, ``Multi-view 3d object detection network for autonomous driving,'' in \textit{Proceedings of the IEEE conference on Computer Vision and Pattern Recognition}, 2017, pp. 1907--1915.

\bibitem{tchuiev2022duqim}
Vladimir Tchuiev, Yakov Miron, Dotan Di Castro, ``Duqim-net: Probabilistic object hierarchy representation for multi-view manipulation,'' in \textit{2022 IEEE/RSJ International Conference on Intelligent Robots and Systems (IROS)}, 2022, pp. 10470--10477.

\bibitem{terry2021pettingzoo}
J K Terry, Benjamin Black, Nathaniel Grammel, Mario Jayakumar, Ananth Hari, Ryan Sullivan, Luis S Santos, Clemens Dieffendahl, Caroline Horsch, Rodrigo Perez-Vicente, and others, ``Pettingzoo: Gym for multi-agent reinforcement learning,'' \textit{Advances in Neural Information Processing Systems}, vol. 34, pp. 15032--15043, 2021.

\bibitem{bansal2018chauffeurnet}
Mayank Bansal, Alex Krizhevsky, Abhijit Ogale, ``Chauffeurnet: Learning to drive by imitating the best and synthesizing the worst,'' \textit{arXiv preprint:1812.03079}, 2018.

\bibitem{kingma2014adam}
Diederik P Kingma, Jimmy Ba, ``Adam: A method for stochastic optimization,'' \textit{arXiv preprint arXiv:1412.6980}, 2014.

\bibitem{AGPNet2}
Yakov Miron, Chana Ross, Yuval Goldfracht, Chen Tessler, Dotan Di Castro, ``Towards Autonomous Grading In The Real World,'' in \textit{2022 IEEE/RSJ International Conference on Intelligent Robots and Systems (IROS)}, 2022, pp. 11940-11946, doi:10.1109/IROS47612.2022.9982114.

\bibitem{AGPNet3}
Yakov Miron, Yuval Goldfracht, Chana Ross, Dotan Di Castro, Itzik Klein, ``Autonomous Dozer Sand Grading Under Localization Uncertainties,'' \textit{IEEE Robotics and Automation Letters}, vol. 8, no. 1, pp. 65-72, 2023, doi:10.1109/LRA.2022.3222990.

\bibitem{SuperSuit}
J K Terry, Benjamin Black, Ananth Hari, ``SuperSuit: Simple Microwrappers for Reinforcement Learning Environments,'' \textit{arXiv preprint arXiv:2008.08932}, 2020.

\bibitem{peng2018sim}
Xue Bin Peng, Marcin Andrychowicz, Wojciech Zaremba, Pieter Abbeel, ``Sim-to-real transfer of robotic control with dynamics randomization,'' in \textit{2018 IEEE international conference on robotics and automation (ICRA)}, 2018, pp. 3803--3810.

\bibitem{pinto2017robust}
Lerrel Pinto, James Davidson, Rahul Sukthankar, Abhinav Gupta, ``Robust adversarial reinforcement learning,'' in \textit{International Conference on Machine Learning}, 2017, pp. 2817--2826.

\bibitem{tessler2019action}
Chen Tessler, Yonathan Efroni, Shie Mannor, ``Action robust reinforcement learning and applications in continuous control,'' in \textit{ICML}, 2019, pp. 6215--6224.

\bibitem{miron2019s}
Yakov Miron, Yona Coscas, ``S-flow gan,'' \textit{arXiv preprint arXiv:1905.08474}, 2019.

\bibitem{rao2020rl}
Kanishka Rao, Chris Harris, Alex Irpan, Sergey Levine, Julian Ibarz, Mohi Khansari, ``Rl-cyclegan: Reinforcement learning aware simulation-to-real,'' in \textit{Proceedings of the IEEE/CVF CVPR}, 2020, pp. 11157--11166.

\bibitem{loquercio2021learning}
Antonio Loquercio, Elia Kaufmann, Ren{\'e} Ranftl, Matthias M{\"u}ller, Vladlen Koltun, Davide Scaramuzza, ``Learning high-speed flight in the wild,'' \textit{Science Robotics}, vol. 6, no. 59, pp. eabg5810, 2021.

\bibitem{sutton2018reinforcement}
Richard S Sutton, Andrew G Barto, ``Reinforcement learning: An introduction,'' MIT press, 2018.

\bibitem{hussein2017imitation}
Ahmed Hussein, Mohamed Medhat Gaber, Eyad Elyan, Chrisina Jayne, ``Imitation learning: A survey of learning methods,'' \textit{ACM Computing Surveys (CSUR)}, vol. 50, no. 2, pp. 1--35, 2017.

\bibitem{ross2010efficient}
St{\'e}phane Ross, Drew Bagnell, ``Efficient reductions for imitation learning,'' in \textit{Proceedings of the thirteenth international conference on artificial intelligence and statistics}, 2010, pp. 661--668.

\bibitem{ross2011reduction}
St{\'e}phane Ross, Geoffrey Gordon, Drew Bagnell, ``A reduction of imitation learning and structured prediction to no-regret online learning,'' in \textit{Proceedings of the fourteenth international conference on artificial intelligence and statistics}, 2011.

\bibitem{vapnik2009new}
Vladimir Vapnik, Akshay Vashist, ``A new learning paradigm: Learning using privileged information,'' \textit{Neural networks}, vol. 22, no. 5-6, pp. 544--557, 2009.

\end{thebibliography}

\end{document}